\documentclass[conference]{IEEEtran}
\IEEEoverridecommandlockouts
\usepackage{cite}
\usepackage{amsmath,amssymb,amsfonts}
\usepackage{algorithmic}
\usepackage{graphicx}
\usepackage{textcomp}
\usepackage{xcolor}
\usepackage{multirow}

\begin{document}
\title{Video-Based Sexual Harassment Detection Using Fine-Tuned Video Transformers}

\author{
\IEEEauthorblockN{Jubayer- Al- Mahmud$^{\dagger}$}
\IEEEauthorblockA{
    \textit{Department of CSE} \\
    \textit{Manarat International University} \\
    Dhaka, Bangladesh \\
    jubayer.am64@gmail.com
}
\and
\IEEEauthorblockN{Md Shamimul Islam$^{\dagger}$}
\IEEEauthorblockA{
    \textit{Department of Computer Science} \\
    \textit{Florida Polytechnic University} \\
    Lakeland, Florida, USA \\
    mislam3051@floridapoly.edu
}
\and
\IEEEauthorblockN{Md Nur Hossain}
\IEEEauthorblockA{
    \textit{Department of CSE} \\
    \textit{Manarat International University} \\
    Dhaka, Bangladesh \\
    nurhossainjoy69@gmail.com
}
\thanks{$^{\dagger}$These authors contributed equally
to this work as first co-authors.
\par\medskip
This work has been submitted to the IEEE for possible publication.
Copyright may be transferred without notice, after which this version
may no longer be accessible.}
}



\maketitle

\begin{abstract}
Sexual harassment represents a pervasive societal challenge, with a substantial proportion of incidents remaining unreported due to insufficient evidence and delayed intervention mechanisms. Smart surveillance systems capable of detecting such incidents in real-time could assist law enforcement and serve as a deterrent. However, video-based harassment detection poses significant challenges, including variations in 
clothing, illumination, partial occlusions, low resolution, and diverse camera viewpoints. Despite its social importance, this problem has received limited research attention, primarily due to the scarcity of large-scale annotated video datasets. To address this gap, we extend the Sexual Harassment Video (SHV) dataset to 600 short video clips sourced from YouTube and annotated into harassment and non-harassment categories using consistent labeling guidelines. We fine-tune three state-of-the-art video transformer models, VideoMAE, ViViT, and TimeSformer, pretrained on Kinetics-400, for binary harassment classification. Experimental results show that all three models achieve competitive classification performance, with TimeSformer attaining the highest test accuracy of \textbf{92.27\%}, outperforming VideoMAE \textbf{91.71\%} and ViViT \textbf{91.16\%} on the extended SHV test set. Our analysis reveals that performance is notably influenced by dataset size, video quality, frame sampling strategy, and model architecture. This work presents a comprehensive evaluation of modern video transformers for this sensitive real-world task.

\end{abstract}


\section{Introduction}
Sexual harassment is one of the most prevalent and harmful forms of misconduct worldwide. According to the United Nations, it encompasses unsolicited sexual advances, requests for sexual favors, and other verbal or physical conduct of a sexual nature~\cite{un_sh,who_sh}, occurring across diverse environments including workplaces, educational institutions, and public spaces~\cite{eeoc_sh}, and manifesting in physical, 
verbal, online, or offline forms~\cite{ilo_sh_definition}. This study focuses on onsite physical harassment detection, as it represents one of the most severe and observable forms of misconduct — without timely intervention, such incidents risk escalating into more serious offenses, underscoring the critical need for automated detection systems.

\begin{figure}[htbp]
\centerline{\includegraphics[width=\columnwidth]{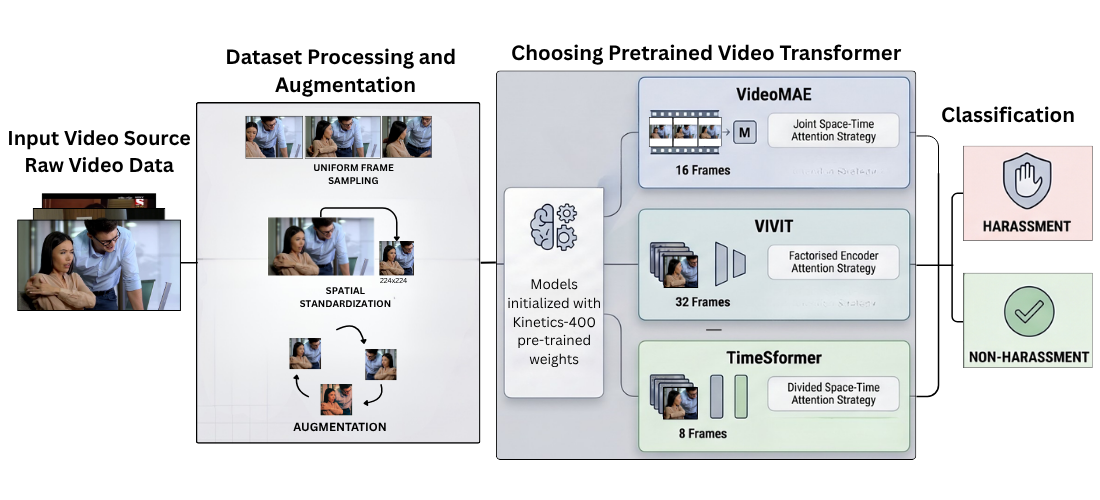}}
\caption{Overview of the data preparation and model initialization 
pipeline for sexual harassment video classification.}
\label{fig:dataset}
\end{figure}

Sexual harassment remains a deeply urgent global concern. The \#MeToo 2024 Report found that over 80\% of women and nearly half of all men report lifetime experiences of sexual harassment or assault~\cite{stopstreetharassment2024}, with nearly 45.1\% of U.S. women experiencing contact sexual 
violence in their lifetimes~\cite{cdc_nisvs2023}. Globally, UN Women highlights that such violence affects individuals across all regions~\cite{unwomen_facts}, underscoring the urgent need for effective automated detection mechanisms.

Recent advances in artificial intelligence offer a promising path toward automated harassment detection. Traditional CNN-based approaches often struggle to capture long-range temporal dependencies in video sequences. In contrast, video transformer architectures leverage self-attention mechanisms 
to model both spatial and temporal relationships across frames, making them particularly well-suited for detecting nuanced behaviors such as harassment. In this work, we fine-tune VideoMAE, ViViT, and TimeSformer on an extended version of the Sexual Harassment Video (SHV) dataset, expanded from 300 to 600 short video clips collected from publicly available YouTube content across both indoor and outdoor environments.

The main contributions of this paper are as follows:
\begin{itemize}
\item We systematically fine-tune VideoMAE, ViViT, and TimeSformer, pretrained on Kinetics-400, for binary sexual harassment classification, demonstrating that TimeSformer achieves the highest accuracy (92.27\%), outperforming VideoMAE (91.71\%) and ViViT (91.16\%), and analyze the influence of frame 
sampling, model architecture, and spatio-temporal attention on detection performance.

\item We expand the SHV dataset from 300 to 600 annotated short video clips, increasing scene diversity across indoor and outdoor environments.

\end{itemize}

The remainder of this paper is organized as follows. 
Section~\ref{sec:related} reviews related work. 
Section~\ref{sec:method} outlines the methodology. 
Section~\ref{sec:results} presents experimental results. 
Section~\ref{sec:conclusion} concludes the paper.

\section{Related Work}
\label{sec:related}

Classifying sexual harassment in videos can be framed as a video action recognition task. Early deep learning approaches relied on spatio-temporal CNNs. Most directly related to our work, \cite{islam2021} proposed a CNN-LSTM framework for video-based sexual harassment detection and introduced the original SHV dataset of 300 video clips, which we extend to 600 clips. Similarly, \cite{roy2022cyberbullying} employed 
deep transfer learning for cyberbullying detection, \cite{zinovyeva2020antisocial} applied NLP-based deep 
learning to identify antisocial online behavior, and \cite{andrade2025susan} introduced SUSAN for detecting 
violence against women in surveillance videos. While these CNN-based methods demonstrated promising results, they are generally limited in modeling long-range temporal dependencies.

The Vision Transformer (ViT)~\cite{dosovitskiy2021} demonstrated that pure transformer architectures can 
effectively model global image context through self-attention, inspiring video transformer models such as 
ViViT~\cite{arnab2021}, TimeSformer~\cite{bertasius2021}, and VideoMAE~\cite{tong2022}. CNN-based methods including I3D~\cite{carreira2017} and X3D~\cite{feichtenhofer2020} often struggle to capture global spatio-temporal context, while hybrid CNN-transformer frameworks such as TransCNN~\cite{ullah2023transcnn} and ViT-based approaches~\cite{tian2024vit} demonstrated that combining pre-trained ViT with spatiotemporal attention can effectively capture local and global relationships in surveillance video.


Several transformer-based works have targeted violence and anomaly detection in surveillance videos. ViViT has been applied to binary violence detection~\cite{singh2022vivit}, ViT with adaptive thresholding achieved near-perfect AUC scores~\cite{crimenet2024}, and attention mechanisms have been leveraged for violence localization~\cite{mohammadi2023video}.Comprehensive surveys~\cite{dilek2025transformers}, semi-supervised approaches~\cite{xing2023svformer}, multi-modal transformers~\cite{chen2021mmvit}, and weakly supervised frameworks~\cite{paulraj2025} further validate transformer effectiveness for surveillance tasks.


Fine-tuning video transformers on domain-specific datasets has shown strong results. Transformer-based approaches have been applied to fine-grained sports action recognition~\cite{lai2024facts}  and transformer-based and video classification models have been applied to gun detection in videos~\cite{videomae_gun}, demonstrating effective generalization to small datasets. Most relevant to our work, \cite{shawon2025} evaluated VideoMAE, ViViT, and TimeSformer on a small-scale domain-specific classification task, showing performance influenced by frame sampling, video quality, and architecture. Fine-tuning for video anomaly detection~\cite{atghaei2023} further validates transfer learning for surveillance tasks.

Despite these advances, no prior work has applied state-of-the-art video transformers specifically to sexual 
harassment detection. This work addresses that gap by fine-tuning VideoMAE, ViViT, and TimeSformer on the extended SHV dataset, providing the first systematic comparison of these architectures for this task.

\section{Methodology}
\label{sec:method}

This study addresses binary harassment classification using the extended SHV dataset. To mitigate challenges posed by limited annotated data and the temporal complexity of human behavior, we fine-tune three pretrained video transformer models to learn discriminative spatio-temporal representations directly from raw video sequences.


\begin{figure}[htbp]
\centerline{\includegraphics[width=\columnwidth]{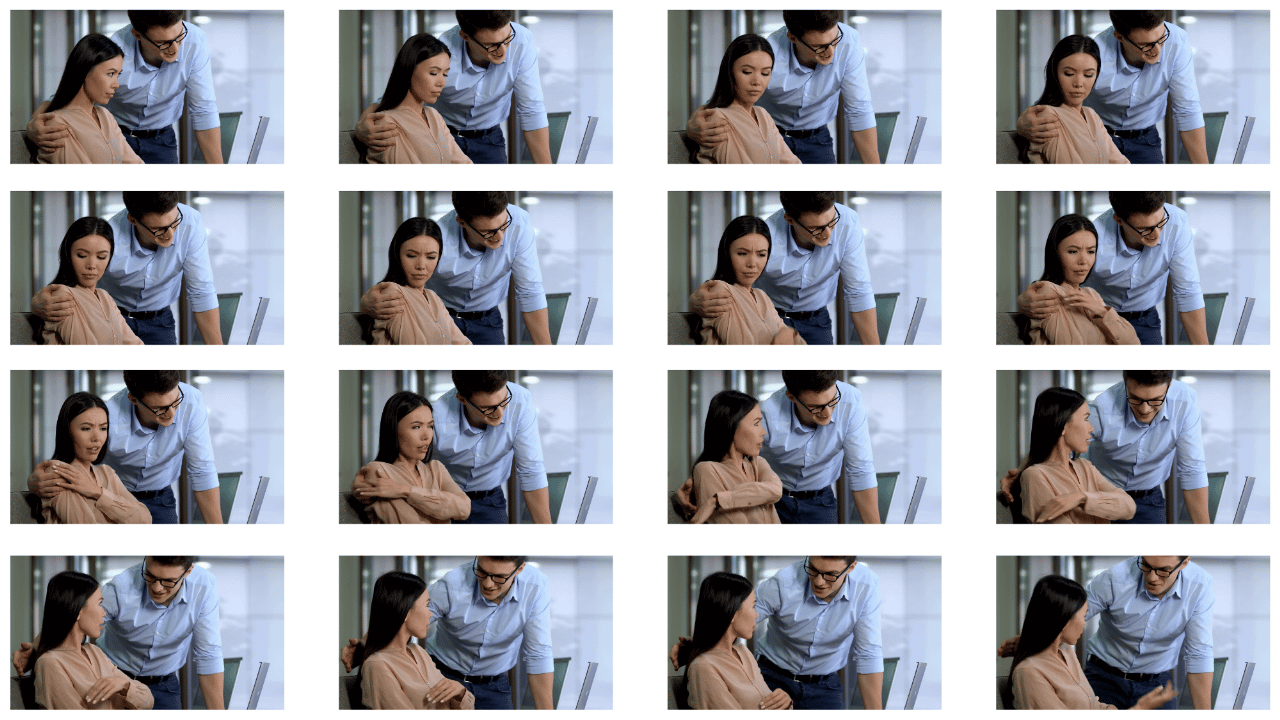}}
\caption{Example of 16 uniformly sampled frames extracted from a harassment video clip in the SHV dataset.}
\label{fig:framesampling}
\end{figure}

\subsection{Dataset and Preprocessing}
The Sexual Harassment Video (SHV) dataset, originally introduced by~\cite{islam2021} with 300 clips, is extended in this work to 600 short video clips collected from publicly available YouTube content covering both indoor and outdoor environments. Harassment videos depict physical sexual harassment such as unwanted touching and molestation, while non-harassment videos capture normal everyday scenarios including walking, office activities, and transportation scenes. The dataset is split into an 80\% training set (197 harassment and 222 non-harassment videos) and a 20\% test set (85 harassment and 96 non-harassment videos), as summarized in Table~\ref{tab:dataset}. Figure~\ref{fig:framesampling} illustrates an example of 16 uniformly sampled frames extracted from a harassment video clip in the SHV dataset.
The 300 newly added clips were extracted from publicly available YouTube videos and manually categorized into harassment and non-harassment classes by the authors during the extraction process, following the same labeling criteria established in the original SHV dataset~\cite{islam2021}.

Non-harassment clips include a variety of friendly and benign physical interactions such as hugging, talking, walking, and other everyday social behaviors, deliberately included to help the models learn to distinguish harassment from superficially similar but non-harassing contact.
Frames are uniformly sampled from each video, extracting 16, 32, and 8 frames for VideoMAE, ViViT, and TimeSformer, respectively, resized to $224\times224$ pixels and normalized using each model's dedicated Hugging Face image processor. For clips shorter than the required frame count, the last frame is repeated to pad the sequence. During training, random resized cropping, horizontal flipping, color jitter, and small random rotation were applied for augmentation. All models were initialized with Kinetics-400 pretrained weights~\cite{carreira2017}, specifically \textit{MCG-NJU/videomae-base-finetuned-kinetics}, \textit{facebook/timesformer-base-finetuned-k400}, and \textit{google/vivit-b-16x2-kinetics400} for VideoMAE, TimeSformer, and ViViT, respectively. Video clips were trained in mini-batches of size 2, with model performance continuously monitored on the test set.

\begin{table}[htbp]
    \centering
    \caption{Sexual Harassment Video (SHV) Dataset Statistics}
    \label{tab:dataset}
    \begin{tabular}{|p{4cm}|p{4cm}|}
        \hline
        \textbf{Description} & \textbf{Details} \\ \hline
        Total Classes & 2 \\ \hline
        Total Videos & 600 \\ \hline
        Harassment Videos & 282 \\ \hline
        Non-Harassment Videos & 318 \\ \hline
        Training Set (Harassment) & 197 \\ \hline
        Training Set (Non-Harassment) & 222 \\ \hline
        Total Training Videos & 419 \\ \hline
        Testing Set (Harassment) & 85 \\ \hline
        Testing Set (Non-Harassment) & 96 \\ \hline
        Total Testing Videos & 181 \\ \hline
        Video Duration & 1--3 seconds \\ \hline
        Preprocessed Resolution & $224 \times 224$ pixels \\ \hline
        Video Source & YouTube \\ \hline
    \end{tabular}
\end{table}




\subsection{Model Configurations}
\label{subsec:modelconfig}


Three transformer-based architectures pretrained on Kinetics-400~\cite{carreira2017} were employed, as 
summarized in Table~\ref{tab:model_config}. Each model adopts a distinct spatio-temporal encoding strategy: 
VideoMAE uses masked autoencoding with tube masking, ViViT applies factorised attention, and TimeSformer 
employs divided space-time attention.

\begin{table*}[htbp]
\centering
\caption{Comparison of video transformer model architectures 
and their configurations used in this study.}
\label{tab:model_config}
\begin{tabular}{|l|p{4.5cm}|p{4.5cm}|p{4.5cm}|}
\hline
\textbf{Configuration} & \textbf{VideoMAE} & \textbf{ViViT} & \textbf{TimeSformer} \\
\hline
Pretrained Checkpoint 
& \textit{MCG-NJU/videomae-base-finetuned-kinetics} 
& \textit{google/vivit-b-16x2-kinetics400} 
& \textit{facebook/timesformer-base-finetuned-k400} \\
\hline
Image Size & 224 & 224 & 224 \\
\hline
Num. Frames & 16 & 32 & 8 \\
\hline
Patch Size & 16 & 16 $\times$ 2 & 16 \\
\hline
Hidden Size & 768 & 768 & 768 \\
\hline
Num. Attention Heads & 12 & 12 & 12 \\
\hline
Num. Hidden Layers & 12 & 12 & 12 \\
\hline
Activation Function 
& GELU & GELU(Fast) & GELU \\
\hline
Trainable Parameters & 86.5M & 86M & 121M \\
\hline
\end{tabular}
\end{table*}

\begin{figure*}[htbp]
\centering
\includegraphics[width=\textwidth]{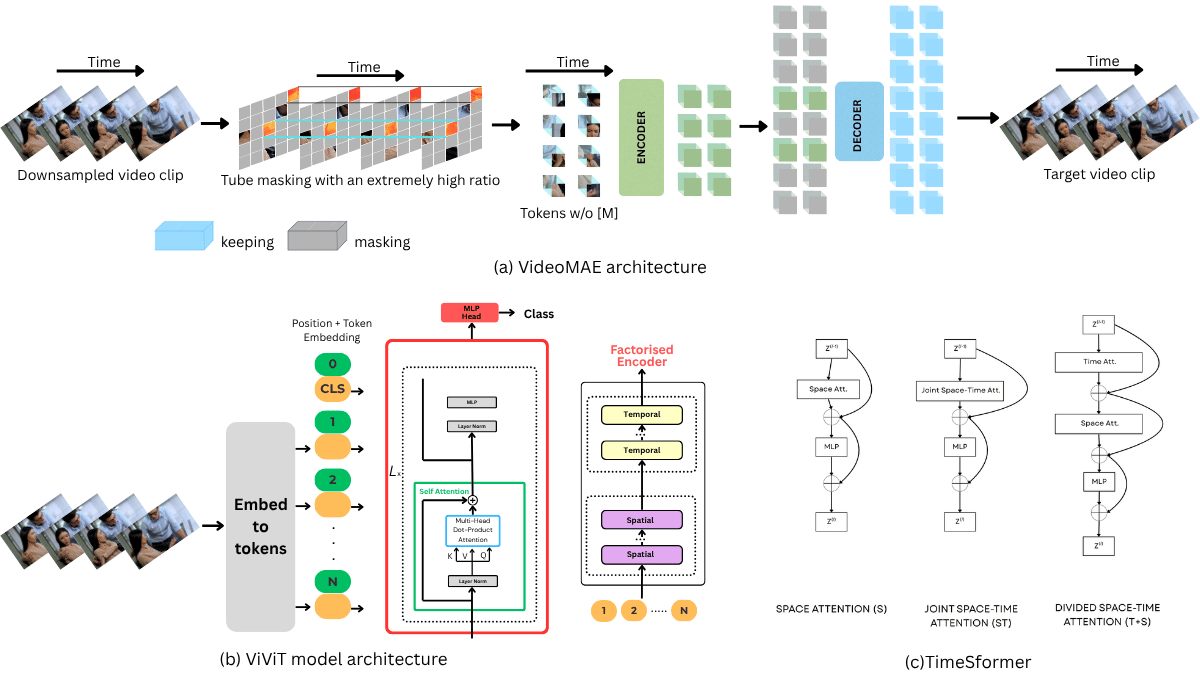}
\caption{Architecture overview of the three video transformer models 
used in this study: (a) VideoMAE, (b) ViViT, and (c) TimeSformer.}
\label{fig:video_transformers}
\end{figure*}

\subsection{Video Masked Autoencoder—VideoMAE Architecture}
\label{subsec:videomae}

VideoMAE~\cite{tong2022} is a self-supervised masked autoencoder for efficient video pre-training. Inspired by Image-MAE~\cite{he2021}, it extends the masked autoencoding paradigm to the video domain using an asymmetric encoder-decoder architecture based on ViT~\cite{dosovitskiy2021}. As illustrated in Figure~\ref{fig:video_transformers}(a), a video clip is temporally downsampled and divided into non-overlapping spatio-temporal cubes of size $2 \times 16 \times 16$ via cube embedding, reducing temporal redundancy while preserving spatio-temporal structure.

To handle high temporal correlation, VideoMAE employs tube masking with a masking ratio of 90\% to 95\%, where entire tubes spanning the same spatial location across consecutive frames are masked together, leaving only 5\% to 10\% of tokens visible to the encoder. This prevents trivial solutions such as frame interpolation or patch tracking, forcing the model to learn richer spatio-temporal representations. The lightweight ViT-based encoder processes only visible tokens, while a small asymmetric decoder reconstructs masked cubes by predicting normalized pixel values using MSE loss, significantly reducing pre-training computational cost.

In this work, VideoMAE is initialized with pretrained weights \textit{MCG-NJU/videomae-base-finetuned-kinetics} from Hugging Face and fine-tuned end-to-end on the SHV dataset. Its strong temporal modeling capability is particularly beneficial for detecting harassment, which often involves subtle aggressive motions, gestures, and behavioral patterns evolving over time. While motion-guided masking in MgMAE~\cite{huang2023} further enhances video masked autoencoding, we adopt the standard VideoMAE 
formulation in our experiments.

\subsection{Video Vision Transformer (ViViT) Architecture}
\label{subsec:vivit}

ViViT~\cite{arnab2021} extends the Vision Transformer (ViT)~\cite{dosovitskiy2021} to video classification by processing spatio-temporal tokens with transformer encoders. Given a video $V \in \mathbb{R}^{T \times H \times W \times C}$, tubelet embedding divides it into non-overlapping 3D tubelets of size $t \times h \times w$, producing $n_t = \frac{T}{t}$, $n_h = \frac{H}{h}$, and $n_w = \frac{W}{w}$ tokens. These are linearly projected, combined with positional encodings, and passed to the transformer encoder.

As shown in Fig.~\ref{fig:video_transformers}(b), ViViT reduces the cost of joint spatio-temporal attention through the following variants:

\begin{enumerate}
\item \textbf{Spatio-Temporal Attention}: Applies attention over all spatio-temporal tokens with complexity $\mathcal{O}((n_t n_h n_w)^2)$.

\item \textbf{Factorised Encoder}: Performs spatial attention followed by temporal attention in separate stages.

\item \textbf{Factorised Self-Attention}: Applies separate spatial and temporal attention blocks within each transformer layer.

\item \textbf{Factorised Dot-Product Attention}: Allocates attention heads to spatial and temporal dimensions and fuses their outputs for improved efficiency.
\end{enumerate}

The model comprises 12 encoder layers with 12 attention heads followed by an MLP classification head. In this work, we fine-tune the Factorised Encoder variant, selected for its balance between computational efficiency and spatio-temporal modeling capacity, using the Kinetics-400 pretrained checkpoint \textit{google/vivit-b-16x2-kinetics400} for binary harassment classification on the SHV dataset. Its factorised attention efficiently captures both spatial appearance and temporal dynamics, making it suitable for detecting subtle aggressive behaviors and interactions over time.

\subsection{TimeSformer Architecture}
\label{subsec:timesformer}

TimeSformer~\cite{bertasius2021} is a pure transformer-based architecture for video understanding that removes convolutions and extends ViT~\cite{dosovitskiy2021} to model spatio-temporal relationships directly from video frames.

As illustrated in Fig.~\ref{fig:video_transformers}(c), a video is divided into non-overlapping patches across frames, which are linearly projected and augmented with learnable positional embeddings. The resulting tokens are processed by a 12-layer transformer encoder with a dedicated classification token.

The key contribution is \textbf{Divided Space-Time Attention}, where each block applies:

\begin{enumerate}
\item \textbf{Temporal Attention}: Each spatial patch attends across frames to capture motion and temporal dynamics.

\item \textbf{Spatial Attention}: Each frame is processed to capture intra-frame spatial context.
\end{enumerate}

This factorization reduces the cost of full spatio-temporal attention while preserving strong modeling capacity. As shown in Fig.~\ref{fig:video_transformers}(c), we use the \textbf{Divided Space-Time Attention (T+S)} variant, which balances efficiency and performance.

The final class token representation is passed to an MLP head for classification. In this work, we initialize TimeSformer with Kinetics-400 pretrained weights \textit{facebook/timesformer-base-finetuned-k400} and fine-tune it end-to-end for binary harassment classification on the SHV dataset. Its strong temporal modeling ability makes it well-suited for detecting subtle aggressive behaviors and evolving interactions in video sequences.

\begin{table*}[t]
\centering
\caption{Per-Class and Average Performance of Fine-Tuned 
Video Transformer Models on the SHV Test Set.}
\label{tab:results}
\begin{tabular}{|p{2cm}|p{2cm}|p{1.1cm}|p{1.1cm}|p{1.1cm}|p{1.1cm}|p{1.1cm}|p{1.1cm}|p{1.1cm}|p{1.1cm}|p{1.1cm}|}\hline
\textbf{Model} & \textbf{Class} & \textbf{Test Loss} 
& \textbf{TP} & \textbf{FP} & \textbf{FN} & \textbf{TN}
& \textbf{Precision} & \textbf{Recall} & \textbf{F1-Score} 
& \textbf{Macro F1} \\
\hline
\multirow{2}{*}{VideoMAE} 
& Harassment     & \multirow{2}{*}{0.2601} & 88 & 8  & 7  & 78 & 0.907 & 0.918 & 0.912 & \multirow{2}{*}{0.917} \\
\cline{2-2}\cline{4-10}
& Non-Harassment &                         & 78 & 7  & 8  & 88 & 0.926 & 0.917 & 0.921 & \\
\hline
\multirow{2}{*}{ViViT}
& Harassment     & \multirow{2}{*}{0.2474} & 93 & 3  & 13 & 72 & 0.960 & 0.847 & 0.900 & \multirow{2}{*}{0.910} \\
\cline{2-2}\cline{4-10}
& Non-Harassment &                         & 72 & 13 & 3  & 93 & 0.877 & 0.969 & 0.921 & \\
\hline
\multirow{2}{*}{TimeSformer}
& Harassment     & \multirow{2}{*}{\textbf{0.2079}} & 94 & 2  & 12 & 73 & 0.973 & 0.859 & 0.912 & \multirow{2}{*}{\textbf{0.922}} \\
\cline{2-2}\cline{4-10}
& Non-Harassment &                                  & 73 & 12 & 2  & 94 & 0.887 & 0.979 & 0.931 & \\
\hline
\end{tabular}
\end{table*}

\begin{figure*}[htbp]
    \centering
    \includegraphics[width=\textwidth]{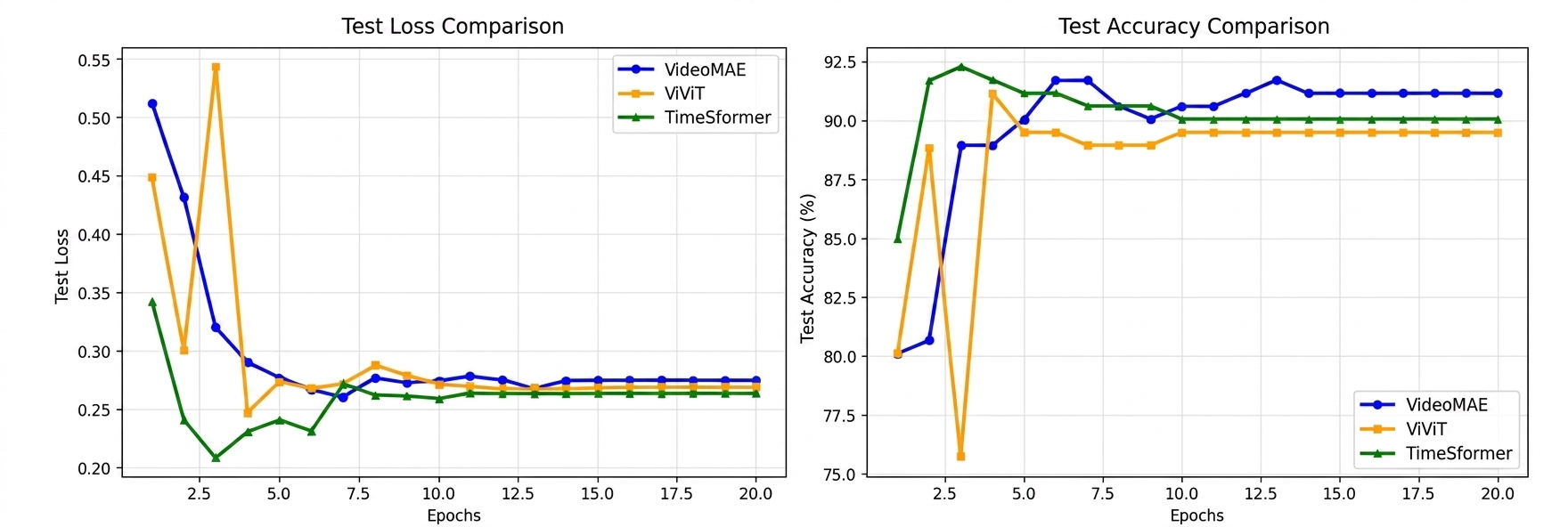}
    \caption{Test loss and accuracy comparison.}
    \label{fig:curves}
\end{figure*}

\subsection{Fine-Tuning of Video Transformers}
\label{subsec:finetuning}

All models are fine-tuned end-to-end for binary harassment
classification. Pretrained Hugging Face weights are loaded
and the final MLP head replaced with a binary output layer.
Training uses AdamW with ReduceLROnPlateau scheduling
and cross-entropy loss. 
All models were trained for up to 20 epochs, 
with test performance monitored throughout to assess convergence and generalization. Full 
hyperparameter details are provided in Table~\ref{tab:hyperparams}.

\begin{table}[htbp]
    \centering
    \caption{Training Hyperparameters for Fine-Tuning}
    \label{tab:hyperparams}
    \begin{tabular}{|p{4cm}|p{4cm}|}
    \hline
        \textbf{Hyperparameter} & \textbf{Value} \\ \hline
        Batch Size & 2 \\ \hline
        Max Epochs & 20 \\ \hline
        Initial Learning Rate & $10^{-5}$ \\ \hline
        Weight Decay & $10^{-4}$ \\ \hline
        Optimizer & AdamW \\ \hline
        Loss Function & Cross-Entropy Loss \\ \hline
        LR Scheduler & ReduceLROnPlateau \\ \hline
    \end{tabular}
\end{table}


\section{Result Analysis}
\label{sec:results}

This section presents the experimental results of the three fine-tuned video transformer models evaluated on the extended SHV test set.

\subsection{Results and Analysis}
The three models are evaluated on the SHV test set (181 videos: 85 harassment, 96 non-harassment) using accuracy, precision, recall, F1-score, and test loss. Table~\ref{tab:results} summarizes the per-class classification performance, confusion matrix results, and average metrics for all three models.

TimeSformer achieves the most stable and highest test accuracy throughout training, the highest non-harassment recall (0.979), and the highest harassment precision (0.973). ViViT exhibits notably higher test loss fluctuation during early epochs (peak test loss $\approx$ 0.55 at epoch 3), attributed to its factorised encoder processing spatial and temporal information in sequential stages. It also achieves the highest harassment precision (0.960) but lower recall (0.847), indicating conservative behavior with higher false negatives. VideoMAE demonstrates smooth and consistent convergence with the most balanced performance, showing similar recall across both classes (0.918 harassment, 0.917 non-harassment).

\subsection{Discussion}
\label{subsec:discussion}

Overall, all three video transformer models achieve strong performance on the SHV task, with test accuracies above 91\%. TimeSformer obtains the highest accuracy (92.27\%) and lowest test loss (0.2079), indicating that its divided space-time attention effectively captures both temporal motion patterns and spatial context, making it well-suited for subtle harassment behavior modeling.

VideoMAE achieves 91.71\% accuracy with the most balanced per-class performance, reflecting robust spatio-temporal representations from masked autoencoding pretraining and making it reliable for real-world surveillance where balanced detection is important. ViViT reaches 91.16\% accuracy and the highest harassment precision (0.960), but lower recall (0.847), indicating a conservative behavior with more missed harassment cases, likely due to its factorised encoder processing spatial and temporal information separately.

Overall, the results confirm that fine-tuning pre-trained video transformers is effective for domain-specific surveillance tasks with limited labeled data, with performance notably influenced by dataset size, video quality, frame sampling strategy, and model architecture~\cite{shawon2025}. Compared to the CNN-LSTM 
baseline~\cite{islam2021} (70\%), all three fine-tuned transformer models demonstrate clear improvement, as summarized in Table~\ref{tab:accuracy}.

\begin{table}[htbp]
    \centering
    \caption{Accuracy comparison on the SHV test set.}
    \label{tab:accuracy}
    \begin{tabular}{|p{4cm}|p{4cm}|}
    \hline
        \textbf{Model} 
        & \textbf{Accuracy (\%)}  \\ 
        \hline
        VideoMAE
        & 91.71 \\
        \hline
        ViViT
        & 91.16 \\
        \hline
        TimeSformer
        & \textbf{92.27} \\
        \hline
        CNN-LSTM \cite{islam2021}
        & 70 \\ 
        \hline
    \end{tabular}
\end{table}

\section{Conclusion and Future Work}
\label{sec:conclusion}

This paper evaluates VideoMAE, ViViT, and TimeSformer for automated sexual harassment detection using an extended SHV dataset expanded from 300 to 600 clips. TimeSformer achieves the highest accuracy (92.27\%), followed by VideoMAE (91.71\%) and ViViT (91.16\%), demonstrating that pretrained video 
transformers effectively capture spatio-temporal harassment patterns even with limited data. Future work will address current limitations including small dataset size and YouTube-sourced bias by expanding to diverse real-world surveillance footage, extending to fine-grained harassment categories, developing lightweight models for real-time deployment, and incorporating multi-modal features such as 
audio, pose information, and temporal localization.

\section*{Data Availability}
The extended SHV dataset (600 clips) used in this study is publicly available on Zenodo at https://doi.org/10.5281/zenodo.22653003.

\end{document}